%% file: root.tex
\documentclass[letterpaper, 10 pt, conference]{ieeeconf}  

\IEEEoverridecommandlockouts                              

\makeatletter
\def\endthebibliography{%
  \def\@noitemerr{\@latex@warning{Empty `thebibliography' environment}}%
  \endlist
}
\makeatother

\usepackage{graphicx} 
\usepackage{epsfig} 
\usepackage{mathptmx} 
\usepackage{times} 
\usepackage{amsmath} 
\usepackage{amssymb}  
\usepackage[caption=false, font=footnotesize]{subfig}
\usepackage{listings}
\usepackage[ruled,linesnumbered]{algorithm2e}
\usepackage{hyperref}

\lstdefinestyle{customc}{
  belowcaptionskip=1\baselineskip,
  breaklines=true,
  frame=L,
  xleftmargin=\parindent,
  language=C,
  showstringspaces=false,
  basicstyle=\footnotesize\ttfamily,
}
\lstdefinestyle{customasm}{
  belowcaptionskip=1\baselineskip,
  frame=L,
  xleftmargin=\parindent,
  language=[x86masm]Assembler,
  basicstyle=\footnotesize\ttfamily,
  commentstyle=\itshape\color{purple!40!black},
}
\title{\LARGE \bf
Pyramidal Blur Aware X-Corner Chessboard Detector
}

\author{Peter Abeles$^{1}$
\thanks{$^{1}$Peter Abeles is with NINOX 360, Redwood City, CA USA {\tt\small pabeles@ninox360.com}}
}

\begin{document}

\maketitle
\thispagestyle{plain}
\pagestyle{plain}

\begin{abstract}
With camera resolution ever increasing and the need to rapidly recalibrate robotic platforms in less than ideal environments, there is a need for faster and more robust chessboard fiducial marker detectors. A new chessboard detector is proposed that is specifically designed for: high resolution images, focus/motion blur, harsh lighting conditions, and background clutter. This is accomplished using a new x-corner detector, where for the first time blur is estimated and used in a novel way to enhance corner localization, edge validation, and connectivity. Performance is measured and compared against other libraries using a diverse set of images created by combining multiple third party datasets and including new specially crafted scenarios designed to stress the state-of-the-art. The proposed detector has the best F1-Score of 0.97, runs 1.9x faster than next fastest, and is a top performer for corner accuracy, while being the only detector to have consistent good performance in all scenarios.
\end{abstract}

\section{Introduction}
Camera calibration is an essential part in determining the mapping between an image's 2D pixel coordinate and the 3D world. By observing calibration markers, it is possible to extrapolate a camera's intrinsic characteristics (e.g. focal length and lens distortion) to a high level of precision \cite{2000_Zhang_Flexible}.

Chessboard (or checkerboard) patterns are arguably the most popular pattern for camera calibration with good reason \cite{2007_Mallon_Pattern}, they are easy to detect and robust to blur due to their symmetry. They are used by themselves or with self-identifying patterns \cite{2008_Fila}, as done in CALTag \cite{2010_CALTag} and ChArUco \cite{2015_Garrido_Charuco}.

In this work, we focus on fully automatic detection of chessboard patterns in diverse environments. Outdoors calibration and in factories is increasingly important and challenging due to poor inconsistent lighting, motion blur, poor focus, harsh shadows, and complex backgrounds. In addition, typical camera resolutions have increased dramatically since much of the past work was proposed and new techniques are needed to handle, the now common, 12+ megapixels images.

A novel chessboard detector is proposed that combines several innovations; a new x-corner detector, blur estimator, robust blur aware edge validation, and overall multi-scale approach. The last three innovations are significant departures from past work and is the first known fiducial detector to estimate the optimal scale to localize a corner, then uses the scale to dynamically sample edges and determine corner connectivity. Enabling heavily blurred and fisheyed images to be processed. The final output can be configured to return all found patterns or use prior information and find a single known target. Results include metadata that is useful for camera diagnostics, autofocus, and estimating corner precision.

To validate performance, a study is performed on what might be the most diverse set of chessboard scenes to date from multiple authors and new ones specifically designed to stress and break detectors in real-world situations. Accuracy is measured using hand labeled corners and synthetic images with known corners. In this study, the proposed detector demonstrated its robustness, accuracy, and speed across all scenarios. Making it the only  tested library to produce consistently good results. Both source code and datasets are freely available online\footnote{Detector and source code has been available since 2019 in BoofCV, with smaller improvements in v0.39. \url{https://boofcv.org}}.

\section{Related Work}

Planar targets are widely used in camera calibration \cite{2000_Zhang_Flexible} for their ease of use and there are a multitude of toolboxes for them, such as OpenCV \cite{2000_OpenCV_Library}, Bouguet's Matlab \cite{2001_Bouguet_Matlab}, CamOdoCal \cite{2013_camodocal}, and BoofCV \cite{2011_BoofCV}. The earliest approaches required human intervention to find corners, while more recent work finds most of the corners automatically, to varying degrees of accuracy and reliability. Deltille patterns have been proposed \cite{2017_Ha_Deltille} as an alternative to chessboards for improved handling of blur and distortion.

One approach to chessboard detection involves using contours around squares. In OpenCV's \textit{findChessboardCorners} \cite{2005_OpenCV} the image is adaptively thresholded, eroded to separate the individual squares/quadrangles, then squares are found using contours. The chessboard grid is formed by connecting quadrangles corners. Image borders are challenging since they crop and/or heavily distort squares. With improvements to this general approach being proposed in \cite{2008_Rufli,2013_camodocal}. Recently, deep learning has been explored for corner detection \cite{2016_Donn_ML}.

Instead of using square contours, x-corners (named after the intersection of two squares) can be detected directly, avoiding image border issues. Harris \cite{1988_Harris} corner detector is the basis for many x-corner detectors \cite{2002_lucchese, 2010_kassir, 2016_Liu}. While Harris is excellent at detecting x-corners, it detects other corner-like shapes too.

Many papers have proposed custom x-corner intensity functions \cite{2012_Geiger,2016_Liu, 2017_Ha_Deltille, 2018_Duda_ChessboardSB}. In \cite{2014_ROCHADE} x-corners are found by looking at binary centerlines \cite{1992_niblack}. Filtering corners after detection is standard. In \cite{2010_kassir, 2012_Geiger} filtering is done using local gradient histogram statistics and based on local intensity patterns in \cite{2016_Liu}.

After candidate corners are found at pixel level precision, they are refined to sub-pixel accuracy \cite{1998_Lavest, 2005_Sun_HighPrice}. Specific approaches include: Fitting a quadratic function \cite{1987_forstner} or saddle point \cite{2002_lucchese, 2014_ROCHADE, 2017_Ha_Deltille} to x-corner intensity image. Maximize gradient orthogonality \cite{2012_Geiger}. Iterative perspective undistort and redetection \cite{2009_datta}.

The first step in constructing chessboard graphs from x-corners is finding high confidence pairs. Combinatorics can be limited by only considering N-nearest-neighbor corners \cite{2010_kassir, 2017_Ha_Deltille, 2018_Duda_ChessboardSB}. Require two corners have compatible orientations/polarities \cite{2012_Geiger,2017_Ha_Deltille}. Reject pairs using image gradient or sample points along edges \cite{2010_kassir, 2010_Dao_Robust, 2017_Ha_Deltille}. 

After pairs are known, the graph is then constructed. This is a challenging process which often drives the final accuracy. In \cite{2012_Geiger,2017_Ha_Deltille,2018_Duda_ChessboardSB} multiple seeds are used then expanded into a graph. In \cite{2013_camodocal} a spline is used to reject rows if the error is too large.

\section{Proposed Approach and Contributions}

The proposed approach is summarized in Algorithms \ref{alg:det_xcorners} and \ref{alg:create_graph}, with a more detailed explanation in the following subsections. "Level" always refers to level in an image pyramid and is synonymous with scale.

X-corner detector contributions:
\begin{enumerate}
\item \textit{Corner localization at optimal scale:} Select lowest level with a strong response to avoid instability.
\item \textit{Explicit handling of x-corner symmetry:} Applying 2x2 box filter for unique local maximum, stabilizing non-maximum suppression, see Figure \ref{fig:xcorner-response}.
\item \textit{Sub-pixel accuracy using mean-shift:} Alternative to curve fitting that is less sensitive to noise.
\end{enumerate}

Graph connectivity and structure contributions:
\begin{enumerate}
\item \textit{Robust edge validation:} Score using $n$-best points. Reduces sensitive to outliers from localized lighting.
\item \textit{Blur aware edge validation:} Avoid sampling poorly defined edges near blurred x-corners centers.
\item \textit{Scale Compatibility:} Only connect a corner to others at same or higher pyramid level. Reduces influence of false positives at lower levels.
\end{enumerate}

\subsection{X-Corner Detector}
\label{sec:xcorner_detection}

\begin{algorithm}
\DontPrintSemicolon
  Let $L$ be the set of images in a pyramid\;
  \For{ $l_i \in L$  } {
	Compute x-corner intensity\;
	2x2 box filter for unique maximums\;
	Non-maximum suppression\;
	Prune using a cascade of filters\;
	Mean-shift sub-pixel localization\;
	Compute corner orientation and revised intensity\;
  }
  Connect corners across pyramid levels\;
  Select level for corner location and orientation\;
\caption{Pyramidal X-Corner Detection\label{alg:det_xcorners}}
\end{algorithm}

\begin{figure}[h]
\begin{center}
\includegraphics[width=0.4\textwidth]{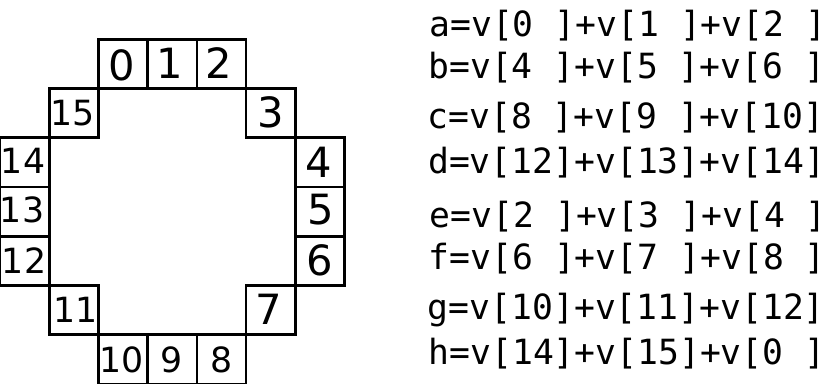}\\
\vspace{0.1cm}
\includegraphics[width=0.4\textwidth]{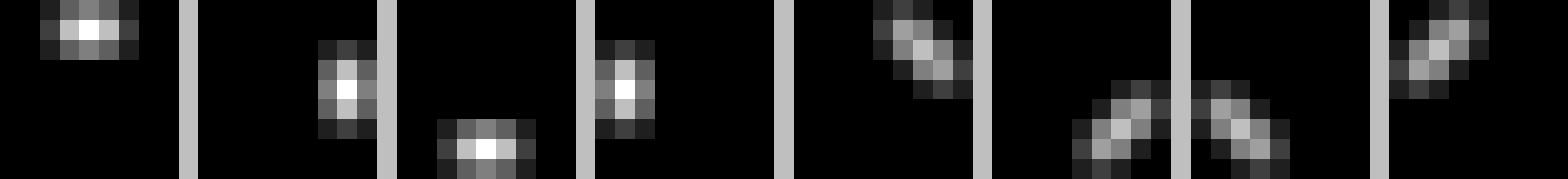}
\caption{Left: Circle sampling pattern stored in array 'v'. Right: Workspace variables formulas. Bottom: Visualization of kernels.}
\label{fig:xcorner-template}
\end{center}
\end{figure}

The proposed x-corner intensity function is designed to be accurate, fast, and have affine lighting invariance. It will be excited only when encountering x-corners. Other corners and lines cause a negative response. The "likelihood" function in \cite{2012_Geiger} is an inspiration, but their formulation has many more logical branches making it expensive to compute.

\begin{figure}[h]
\centering
\subfloat[input]{\includegraphics[width=0.12\textwidth]{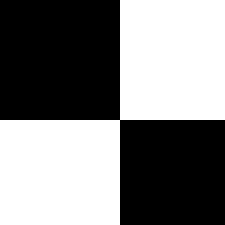}} \qquad
\subfloat[x-corner]{\includegraphics[width=0.12\textwidth]{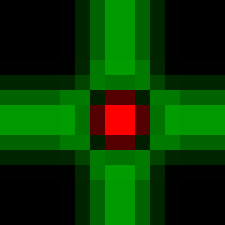}} \qquad
\subfloat[filter 2x2]{\includegraphics[width=0.12\textwidth]{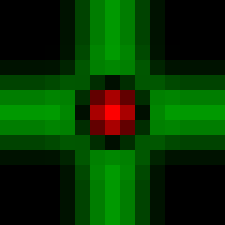}}
\caption{a) Chessboard corner without perspective distortion. b) Original x-corner intensity. Red and green indicate positive and negative x-corner response, respectively. c) Local maximum after applying 2x2 block filter, which improves corner localization in general.}
\label{fig:xcorner-response}
\end{figure}

\emph{Corner Intensity:} 1) Convert image to grayscale and scale pixel values to be 0 to 1. 2) Apply 3x3 Gaussian blur to image. 3) For each pixel, sample a circle as shown in Figure \ref{fig:xcorner-template}. 4) Compute x-corner intensity $I \in \Re$
\begin{equation}
\label{eq:xscore-inten}
I = \mbox{max}\left(\mbox{xscore}(a,b,c,d),\mbox{xscore}(e,f,g,h)\right)
\end{equation}
where $\{a,b,c,d,e,f,g,h\}$ are sampled in Step 3.
\begin{lstlisting}[mathescape=true]
fun xscore( $v_1$ , $v_2$ , $v_3$ , $v_4$)
  $\mu$ = ($v_1$+$v_2$+$v_3$+$v_4$)/4
  return ($v_1$-$\mu$)*($v_3$-$\mu$) + ($v_2$-$\mu$)*($v_4$-$\mu$)
\end{lstlisting} 
Gaussian blur in Step 2 acts as a computationally efficient template to construct kernels in Figure \ref{fig:xcorner-template}. The \emph{xscore} function is at a maximum when the kernel response $v_1$ and $v_3$ are both above or below the mean, then the same is done for $v_2$ and $v_4$.  Affine lighting invariance is achieved by subtracting the mean then multiplying the difference. Partial rotation invariance is achieved in Eq. \ref{eq:xscore-inten} by using two templates offset by 45 degrees. 

In the absence of noise and perspective distortion, Eq. \ref{eq:xscore-inten} will not have unique local maximums. This is resolved by convolving a 2x2 box kernel to the intensity image (Figure \ref{fig:xcorner-response}) and apply (0.5, 0.5) pixels offset to undo box filter induced shift. In typical situations, this helps sub-pixel estimation converge by providing a better initial estimate. For corners at a skewed angle and heavy fisheye distortion the box filter still help break the symmetry.

A cascade of filters is applied to reduce candidate corners in order of least to most computationally expensive. 1) Filter corners based on intensity relative to maximum intensity in top pyramid level. 2) Filter if too many positive x-corner intensity values. Actual x-corners tend to have negative neighbors. 3) Check for expected up-down gray scale pattern. 4) Apply Shi-Tomasi \cite{1994_Shi_GoodFeatures} corner Eigenvalue test.

\begin{figure}[h]
\begin{center}
\includegraphics[width=0.4\textwidth]{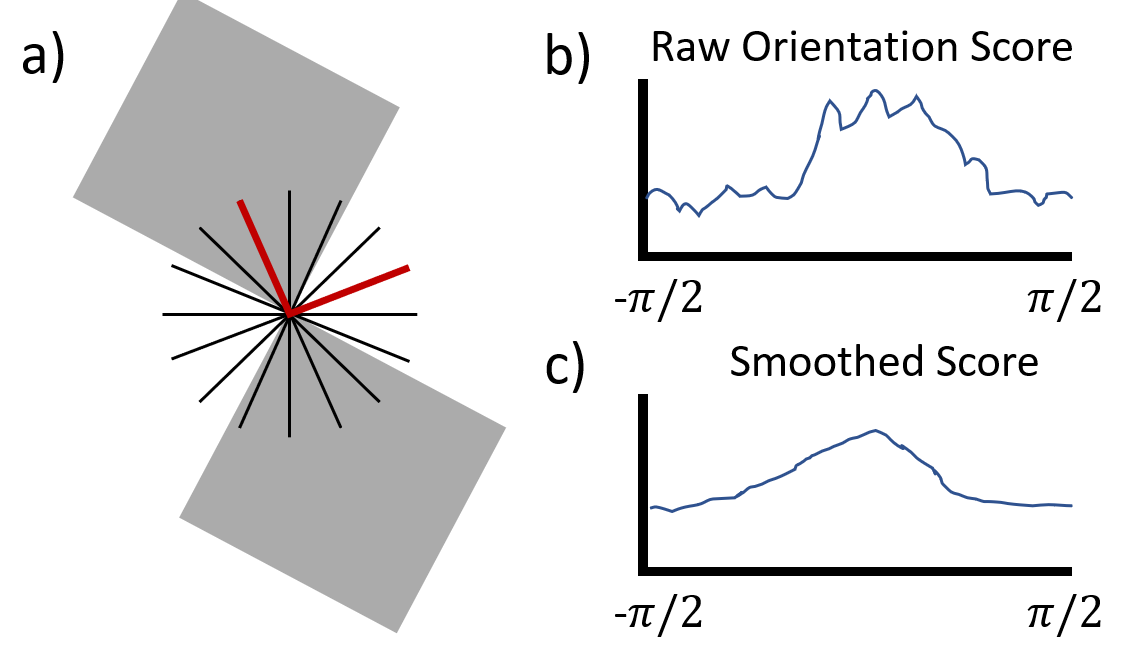}
\caption{a) Orientation and a rotationally invariant x-corner intensity are found using the line integral along 32 spokes centered on the corner. Red indicates selected orientation. b) Cartoon showing raw spoke orientation score. c) Unique maximum after smoothing.}
\label{fig:spokes}
\end{center}
\end{figure}

Remaining corners are refined to sub-pixel accuracy using mean-shift in the intensity image. Mean-shift implicitly applies a low pass filter, making it less sensitive to noise than curve fitting approaches, which attempt to exactly fit all values. Orientation and an improved x-corner intensity are estimated using spokes, Figure \ref{fig:spokes}. These spokes are conceptually similar to the approximated Radeon transform in \cite{2018_Duda_ChessboardSB}. Corner half-circle orientation is estimated by computing the difference between each spoke and one offset by 90 degrees. Resulting array is smoothed with a Gaussian kernel to create a maximum in the middle, Figure \ref{fig:spokes}. The selected angle is further refined by fitting a quadratic. The new x-corner intensity is set to the orientation's best score and exhibits better rotational invariance than Eq \ref{eq:xscore-inten}. Corner contrast is computed as the difference in magnitude of spokes for the best orientation and is used later on. 

\subsection{Pyramidal Processing}

\begin{figure}[h]
\centering
\subfloat[Level 0]{\includegraphics[width=0.15\textwidth]{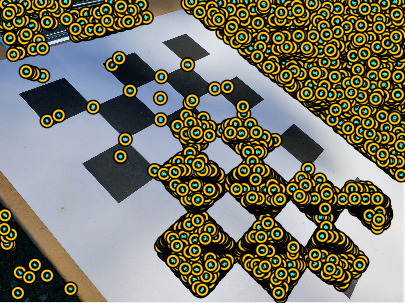}}
\subfloat[Level 1]{\includegraphics[width=0.15\textwidth]{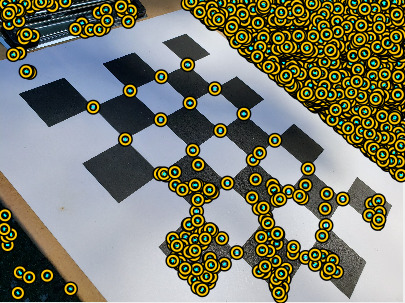}}
\subfloat[Level 2]{\includegraphics[width=0.15\textwidth]{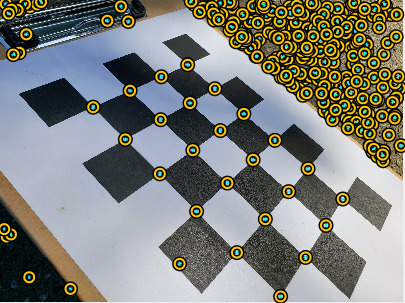}}
\\
\subfloat[Level 3]{\includegraphics[width=0.15\textwidth]{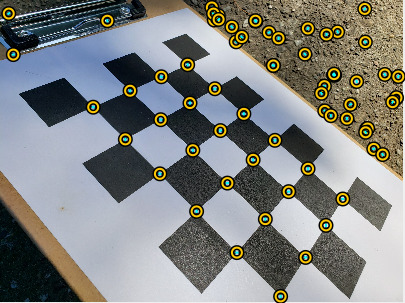}}
\subfloat[Level 4]{\includegraphics[width=0.15\textwidth]{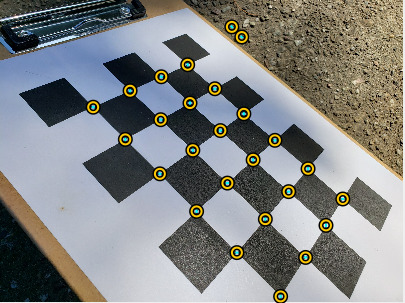}}
\caption{Corners detected across pyramid levels. Higher levels have lower resolution. Yellow circles are found x-corners.}
\label{fig:pyramid-processing}
\end{figure}

Unlike blob features \cite{1999_Lowe_SIFT}, e.g. Hessian or Laplacian, corners do not have a unique maximum in intensity across scale-space \cite{1987_Witkin_ScaleSpace} but have a constant intensity. This is similar to the aperture problem \cite{1994_Shi} but across scale-space. In fact, if there is no noise or blur, a single corner would be observable in all pyramid levels. In practice, the first level is determined by the amount of blur and sensor noise, while the last level is based on ratio of marker size and distance.

Corners are poorly defined under heavy blur and localization is dominated by noise. Naively assuming higher resolution are better for localization causes instability. To overcome this issue, we proposed that the selected level for localization maximizes resolution and intensity:
\begin{equation}
\max_{level} \quad \mbox{intensity}(c,level)/(level+1)
\end{equation}
where $\mbox{intensity}(c,level)$ is a corner's intensity at $level$.

Metadata from pyramidal processing is useful for end consumers. High confidence corners will typically be viewed across multiple levels. The first observed level indicates the amount of local blur/noise. For example, this could be used to autofocus a camera on chessboard patterns, estimate corner accuracy, or detect faults during calibration.

\subsection{Corner Connectivity}

\begin{algorithm}
\DontPrintSemicolon
Let $L$ be the set of all pyramid levels\;
\For{ $l_i \in L$ }{
Let $C$ be the set of all detected corners in $l_i$ \;
\For{ $c_i \in C$ } {
Let $C_n$ be the set of nearest neighbors of $c_i$ visible in the same level or greater\;
\For{ $c_j \in C_n$ } {
	Discard $c_j$ if incompatible orientation\;
	Sample grid of points connecting $c_i$ and $c_j$\;
	Score samples using binary intensities\;
	Connection if sufficient score\; 
}
}
}
\For{ $c_i \in C$ } {
	Vote on connected neighbors\;
	Select neighbors based on perspective geometry and graph constraints\;
}
Discard ambiguous corners based on votes\;
Sort and enforce graph constraints\;
\caption{Chessboard Graph Formation\label{alg:create_graph}}
\end{algorithm}

\begin{figure}[h]
\centering
\includegraphics[width=0.4\textwidth]{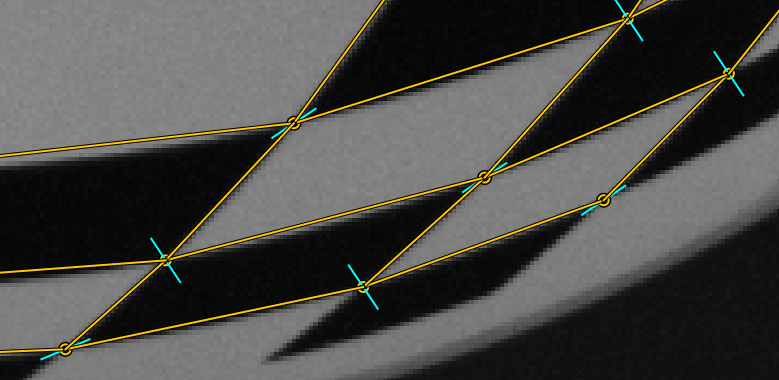}
\caption{Visualization of corner orientation (cyan) and corner connections (orange) under fisheye distortion}
\label{fig:fisheye-connect}
\end{figure}

Connected corner pairs are required to construct a topological graph that represents the chessboard. These connections have known topological properties and orientations. Procedure: 1) For each corner, find its $k$-nearest-neighbors (KNN) in the same pyramid level or above as candidates for connecting. 2) Reject if the corner orientations are not perpendicular. 3) Reject if edge validation fails.

\begin{figure}[h]
\centering
\includegraphics[width=0.45\textwidth]{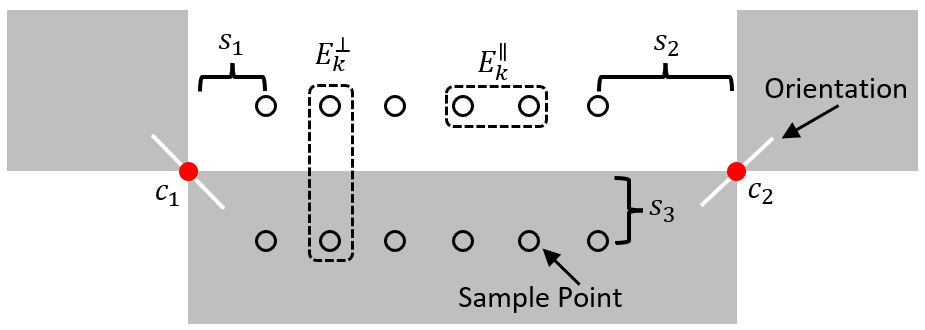}
\caption{Diagram illustrating components of edge validation. Sample locations are dynamic based on corner level to avoid blur.}
\label{fig:edge_sampling}
\end{figure}

Edge validation is done by sampling points in a grid on either side of the line, Figure \ref{fig:edge_sampling}. $\{c_1, c_2\}$ are corners with different amounts of local blur. Spacing $s_i$ is dependent on the pyramid level of $c_i$. The proposed novel dynamic spacing based on level avoids sampling poorly defined edges near blurry corners. Perpendicular offset $s_3$ is determined by the distance between $c_1$ and $c_2$. Perpendicular error $E^{\bot} = I_i - I_j$ score is found by sampling adjacent points across the edge and is maximized by high contrast. Longitudinal error $E^{\parallel} = |I_i - I_j|$ measures similarity. These scores are sorted and the worst values removed to prevent outliers caused by noise and hard shadows from dominating. The edge intensity score is shown below and edges with values below a threshold are pruned:
\begin{equation}
L_{ij} = \frac{\sum_k E^{\bot}_k - E^{\parallel}_k}{\mbox{contrast}(c_i) + \mbox{contrast}(c_j)}
\end{equation}
Corner $\mbox{contrast}(c)$ is defined in Section \ref{sec:xcorner_detection} and is used as a divisor for lighting invariance.

\subsection{Grid Graph Construction}
\label{sec:graph-construction}

A correctly formed chessboard topological graph will have the following properties: 1) All corners must be mutually connected. 2) Every corner must have 2, 3, or 4 neighbors. 3) There must be one and only one other common corner between two neighbors that are adjacent. Any connection or corner not meeting these conditions is pruned. If there are too many connections and ambiguous corners have been identified, then a vote is taken from the local neighborhood where each corner decides independently which connection is the best fit based on edge intensity and expected geometry. The losing corner(s) are removed.

\subsection{Grid Graph to Chessboard Graph}

Grid graphs are reformatted into valid chessboard graphs by putting corners into a canonical counter-clock-wise graph order using corner orientation information and ensuring corner (0,0) is connected to a corner square. Optionally, a single complete grid can be enforced by removing stray connections: Outer rows and columns are removed if there are missing corners. This is repeated as required. If the chessboard shape is known in advance then only graphs which match are returned. If only a single chessboard is expected then the pattern with the largest apparent image size is returned. In summary, output can be multiple arbitrary chessboards or a single known chessboard where false positives are pruned with stronger assumptions.

Please consult source code for specifics as algorithm details in this section have been simplified for conciseness.

\section{Performance Study}

\captionsetup[subfigure]{labelformat=empty}
\begin{figure}[h]
\subfloat[Border]{\includegraphics[width=0.200\linewidth]{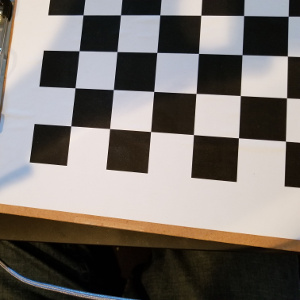}} \hfill
\subfloat[Caltech Edges]{\includegraphics[width=0.200\linewidth]{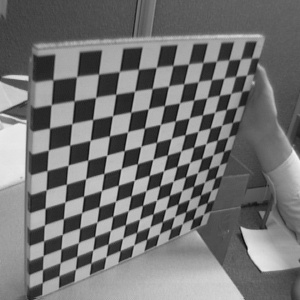}} \hfill
\subfloat[Distant]{\includegraphics[width=0.200\linewidth]{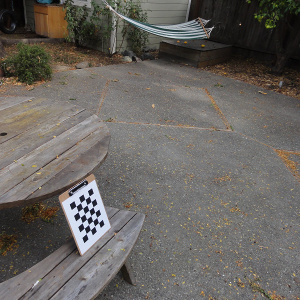}} \hfill
\subfloat[Gaussian]{\includegraphics[width=0.200\linewidth]{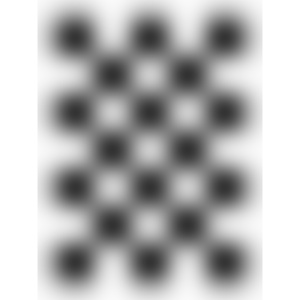}} \\
\subfloat[Large Shadow]{\includegraphics[width=0.200\linewidth]{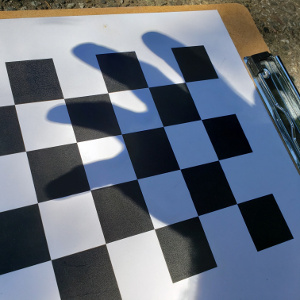}} \hfill
\subfloat[OCam Fisheye190]{\includegraphics[width=0.200\linewidth]{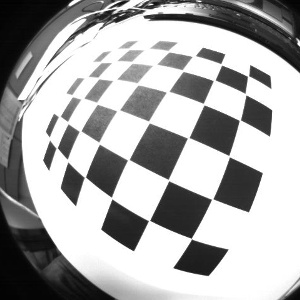}} \hfill
\subfloat[Perfect]{\includegraphics[width=0.200\linewidth]{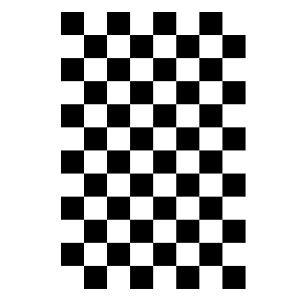}} \hfill
\subfloat[Rotation Vertical]{\includegraphics[width=0.200\linewidth]{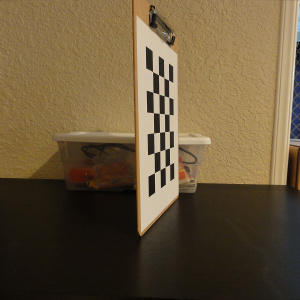}} \hfill
\caption{Select sample images from 8 of 24 the scenarios. A wide range of environments, lighting conditions, and sensors have been tested.}
\label{fig:datasets}
\end{figure}

A diverse performance study is conducted by combining several datasets from multiple authors, as well as adding new scenarios specifically designed to stress detectors, Figure \ref{fig:datasets}. Scenarios vary widely in camera type, image size (0.3 MP to 12 MP), lighting conditions, noise level, motion blur, focus, and background clutter. Corner accuracy is measured using hand labeled corners as well as synthetic images with known corner locations. Every image will have one and only chessboard pattern with all corners visible, as is typical for single camera calibration. See technical report \cite{2021_Abeles_Techreport} for a more complete description of all scenarios.

Performance is measured for chessboard detection, corner localization accuracy, and runtime performance. Detection accuracy is measured using F1-Score statistics. A \emph{true positive} TP is defined as a detected chessboard where every corner is within $t_c$ pixels of ground truth. A \emph{false positive} FP is defined as a detection with the expected grid shape and one or more corners outside of $t_c$. Detections with the wrong shape or missing corners are ignored. To keep it simple, because chessboard orientation can be ambiguous, corner error is computed using the closest match in ground truth. If possible, the known target's shape is given to a library.

Corner labels are used instead of the common post calibration reprojection error because reprojection error does not measure the detector's accuracy directly. Instead reprojection error measures the calibration system as a whole and can hide systematic bias and report optimistic results due to camera model over parameterization. As an example, if a chessboard detector is biased by 5 pixels the calibration system would compensate by translating the camera and report perfect results.

Hand labeled x-corners are estimated to have an accuracy of around 0.2 pixels when images are in focus. Labeling heavily blurred images is difficult to impossible. For this reason, the \emph{Gaussian} scenario is simulated with exact ground truth and \emph{Gaussian Fisheye} has Gaussian blur added to a real fisheye image. By default, $t_c$ is set to a generous 5 pixels and for \emph{focus\_large} $t_c=20$ due to labeling accuracy. The relative rankings of each library is insensitive to the choice of $t_c$, Figure \ref{fig:fscore-overall}. Libraries which define the pixel at $(1,1)$ to be within $x,y \in (0.5,1.5]$ are offset by (0.5, 0.5) pixels.

Runtime is measured using system clock in code while excluding IO. Libraries written in C/C++ are compiled in release mode and built with GCC 9.3.0. Java libraries use JVM 15.0.3 and are given a warm start. Matlab libraries run in headless mode using version R2021a.

Libraries and Versions: \emph{Proposed:} \cite{2011_BoofCV} BoofCV v0.39, \emph{DelChe:} \cite{2017_Ha_Deltille} SHA eed7e86, \emph{Geiger:} \cite{2012_Geiger} libomnical and libcbdetect from website 2021-Aug-15, \emph{OCamCalib:} \cite{2006_OCamCalib} v3.0, \emph{Binary:} OpenCV 4.5.3 findChessboardCorners and cornerSubPix, \emph{ChessSB:} \cite{2018_Duda_ChessboardSB} OpenCV 4.5.3 findChessboardCornersSB. To be clear, \emph{DelChe} is the chessboard detector which was adapted from a deltille pattern detector.

External Datasets: \emph{caltech\_edges} \cite{2001_Bouguet_Matlab}, \emph{ocam} \cite{2006_OCamCalib}, \emph{stefano\_2012} \cite{2006_Stefano_Images}, \emph{challenge} \cite{2010_Peynot_Challenge}, \emph{kaist fisheye} \cite{2017_Ha_Deltille}.

\section{Results}

Corner localization accuracy is reported using 50\% error $E50$ and max error statistics $E100$. The same is true for runtime performance, i.e. $R50$ and $R100$. Median or 50\% statistics is intended to represent "expected" or nominal performance with outliers removed. Max statistics is the worst case performance and tends to be volatile. $N$ is the number of images in a specific scenario. See Tech Report \cite{2021_Abeles_Techreport} for a more complete discussion of these results.

Libraries which crashed, froze, or were excessively slow presented a challenge for summary results. Due to the lack of any better options, summary performance is computed by omitting scenarios a library could not process. This gives some libraries a significant advantage by removing the most difficult scenarios. Here is a list of libraries and the number of omitted scenarios; DelChe 2, and OCamCalib 1.

\begin{figure}[h]
\centering
\includegraphics[width=.5\textwidth]{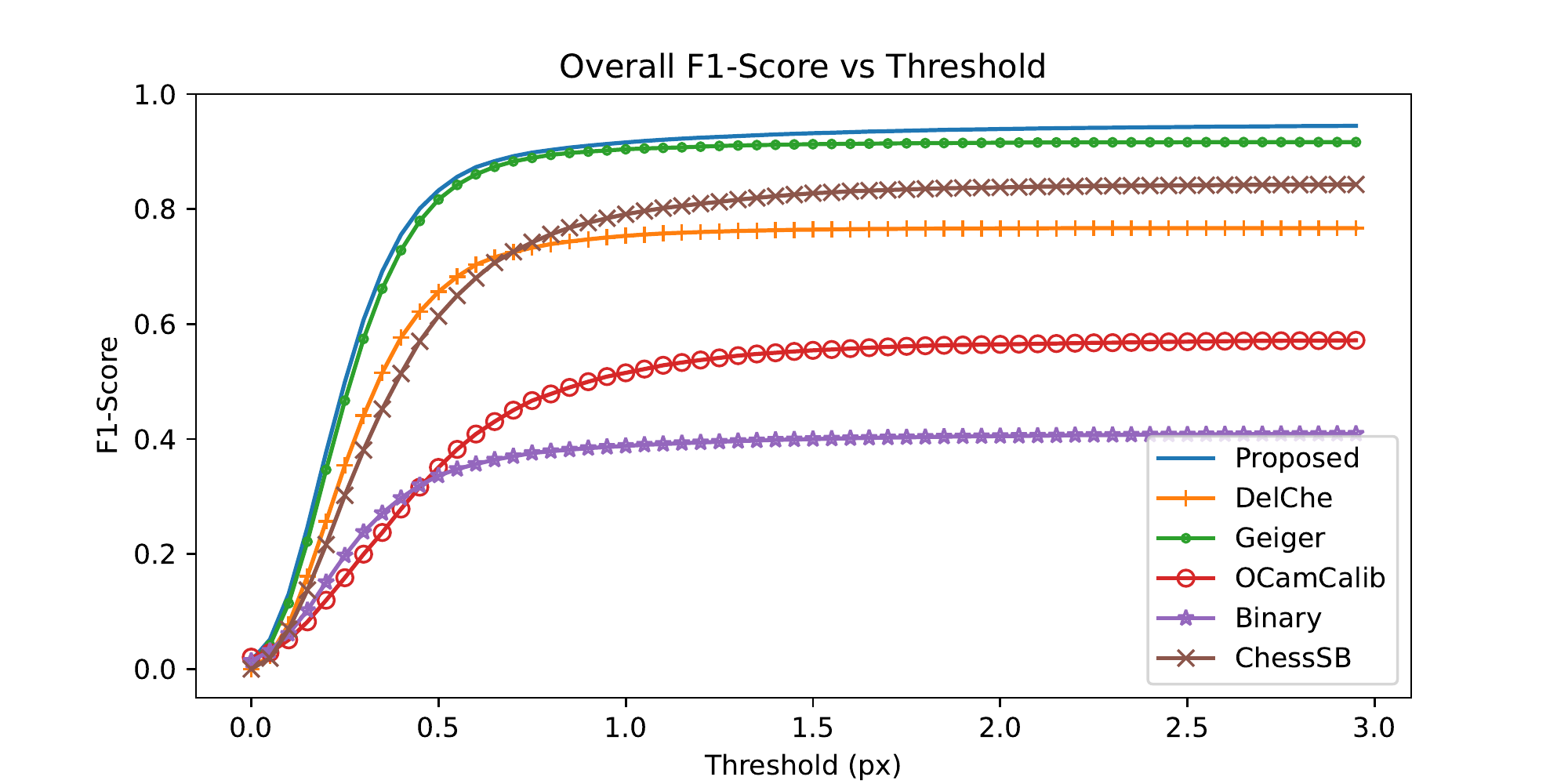}
\caption{F1-Score by combining all scenarios with different thresholds}
\label{fig:fscore-overall}
\end{figure}

\begin{figure}[h]
\centering
\includegraphics[width=.45\textwidth]{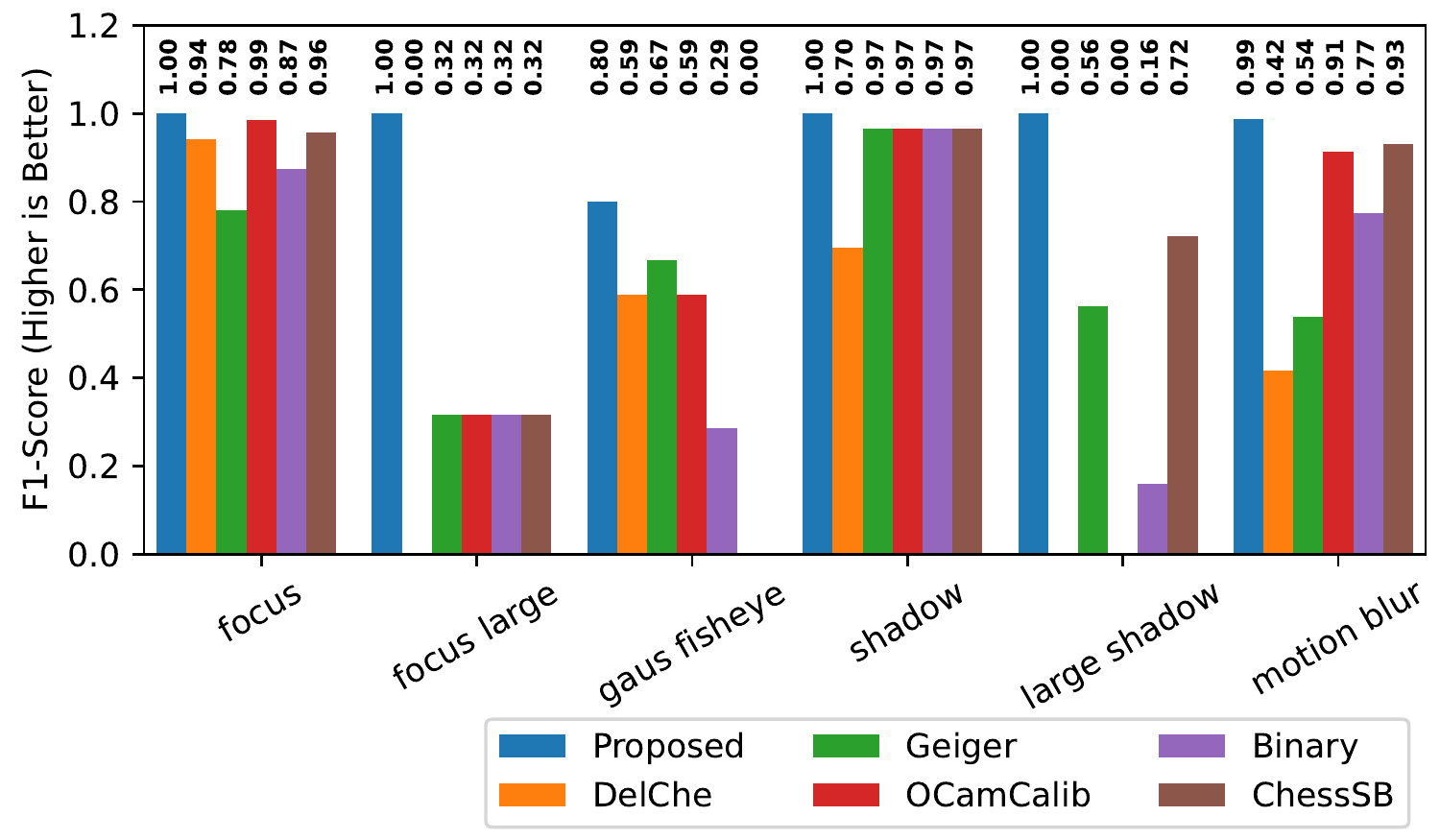}
\caption{F1-Score in select challenging scenarios}
\label{fig:fscore-degraded}
\end{figure}

Detection performance as a function of true positive thresholds across all scenarios combined is shown in Figure \ref{fig:fscore-overall}. The proposed approach has the best performance with an F1-score of 0.97 using a 5 pixel threshold. The next best library is Geiger with an F1-score of 0.92. The proposed and Geiger have comparable performance in nominal situations, but the proposed approach outperforms Geiger in degraded scenarios. Figure \ref{fig:fscore-degraded} shows performance in select scenarios with blur or adverse environmental conditions. In these individual scenarios the proposed is the top performer, often by a considerable margin, i.e. 3x in "focus large". The biggest performance difference tends to be in scenarios with larger images ($\geq 10$ megapixels, see Table \ref{table:results_proposed}), validating the proposed blur aware pyramidal approach.

\begin{figure}[h]
\centering
\includegraphics[width=.38\textwidth]{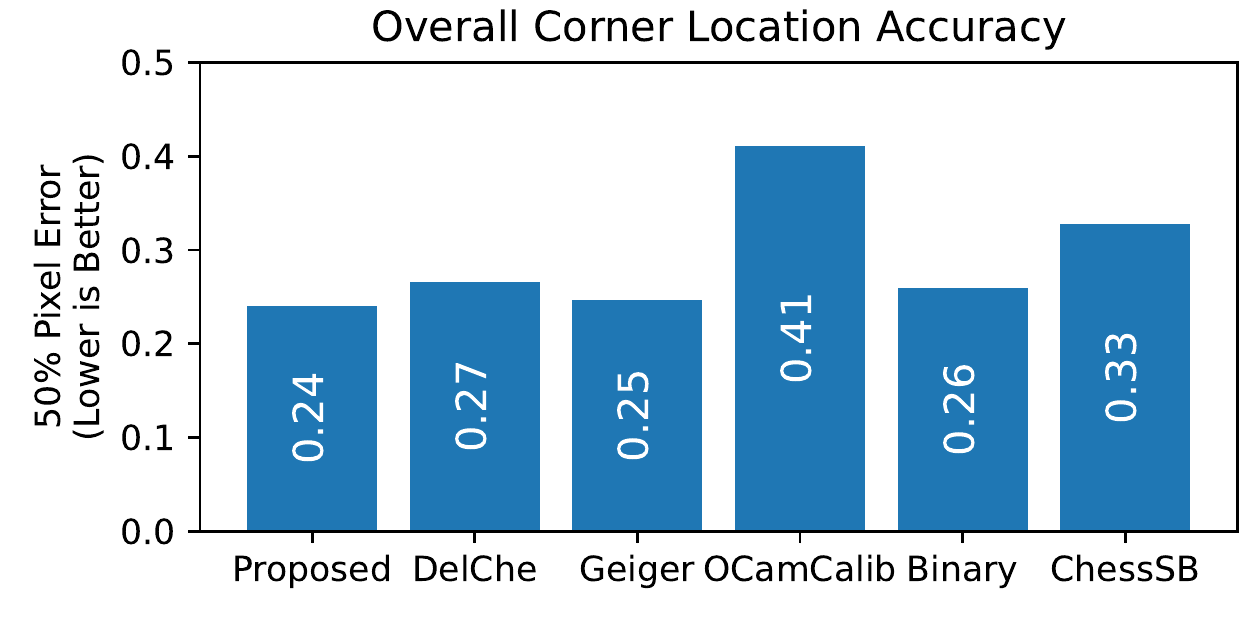}
\caption{Overall Corner Location Accuracy}
\label{fig:error-overall}
\end{figure}

Figure \ref{fig:error-overall} shows nominal corner localization accuracy using 50\% error from all true positive detections across all scenarios. Using 50\% error avoids putting libraries with more detections at a disadvantage by excluding outliers. The proposed detector is effectively tied for best nominal accuracy. All libraries exhibit reasonable nominal accuracy.

\begin{table}[ht]
\setlength{\tabcolsep}{4pt}
\centering
\input{accuracy_table.tex}
\caption{Comparison of Corner Accuracy in Select Scenarios}
\label{table:accuracy_table}
\end{table}

Because overall nominal performance paints an unrealistically optimistic picture, selected specific scenarios are now considered. Table \ref{table:accuracy_table} shows true positives, 50\% corner error, and max corner error in scenarios where most detectors exhibited "good" detection rates. \textit{Sloppy} is a printed on wavy paper, \textit{Motion} has motion blur, \textit{Stef2012} regular camera, and \textit{ThetaS} is an 185$^{\circ}$ fisheye. The reason max error tops out around 5 pixels is because detentions are labeled as false positives at that point. Even if a detector has very low 50\% error accuracy, calibration results will be poor if the max and false positive rates are high.

\begin{figure}[h]
\centering
\includegraphics[trim=40 40 25 40,width=.22\textwidth]{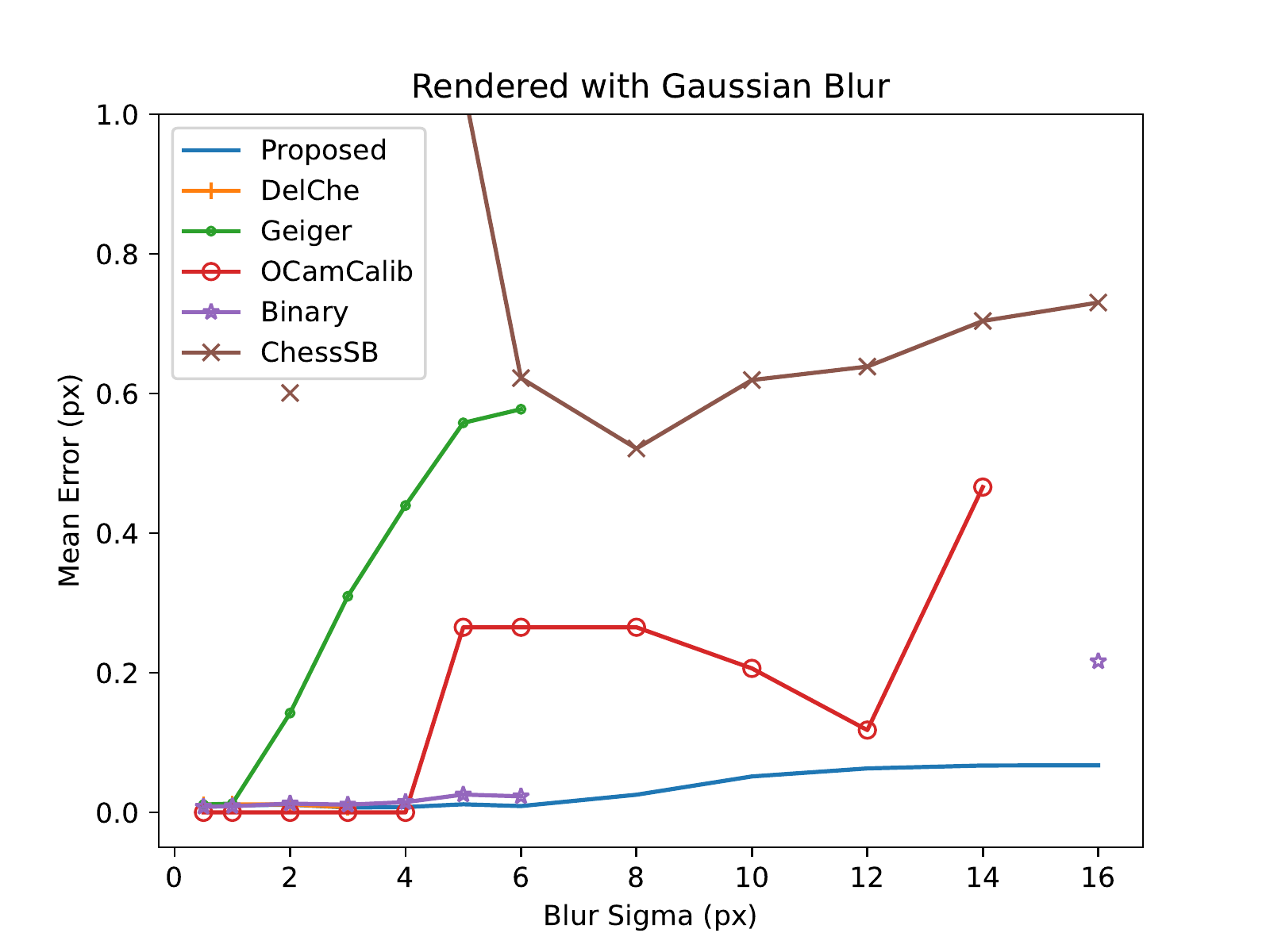}
\includegraphics[trim=40 40 25 40,width=.22\textwidth]{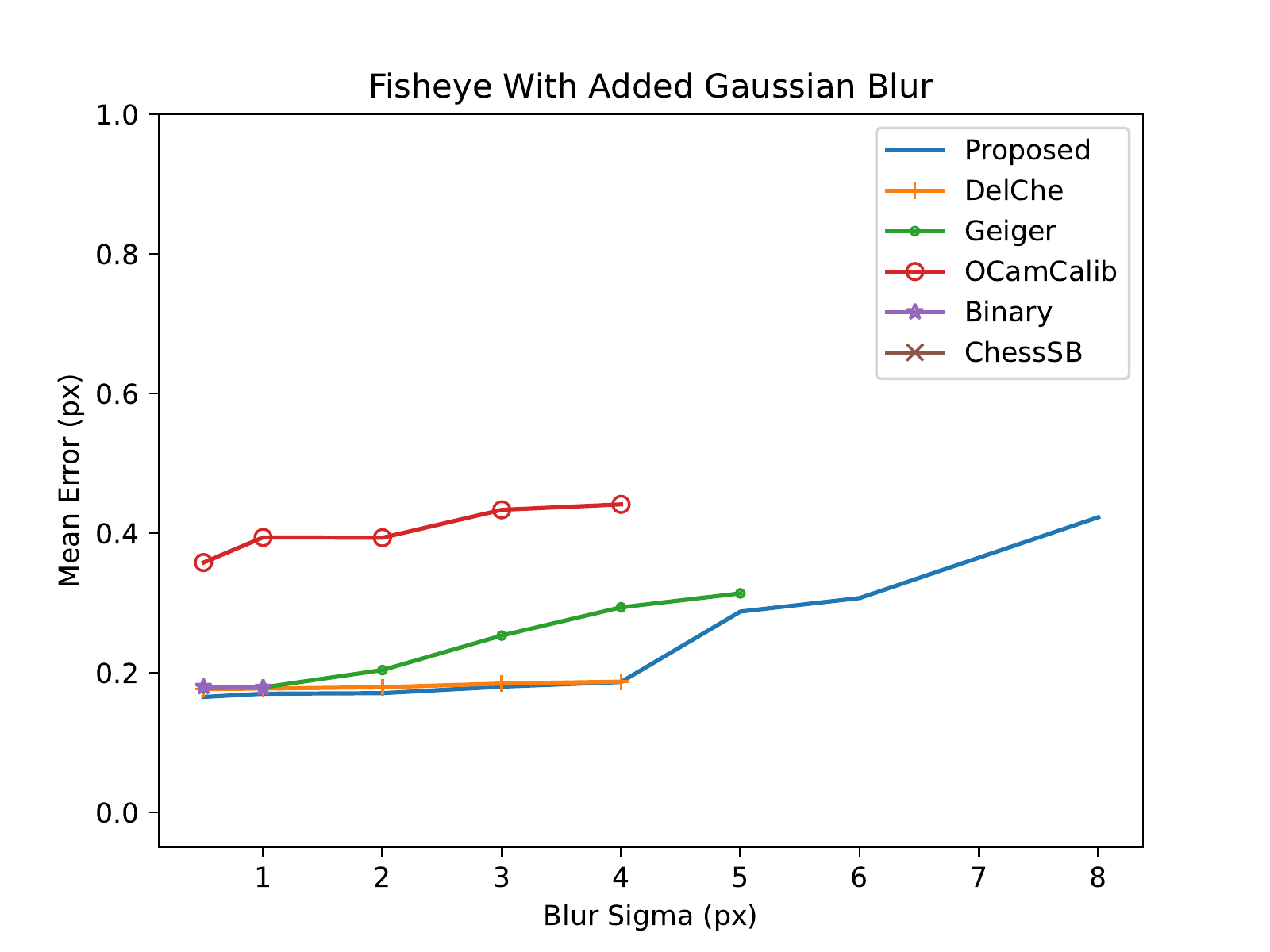}
\caption{Accuracy plots with increasing Gaussian blur. Left: Rendered chessboard without perspective distortion. Right: Fisheye with hand labeled corners.}
\label{fig:accuracy-gaus-perfect}
\end{figure}

Figure \ref{fig:accuracy-gaus-perfect} shows accuracy degrading as a function of Gaussian blur in two scenarios. Ideally a detector would slowly degrade as blur increases and spikes could indicate fundamental instability in the approach. Oddly enough, the rendered scenario with no lens or perspective distortion is very challenging for some detectors, while the fisheye scenario exhibits stable performance. This could be caused by x-corner symmetry discussed previously. In both scenarios, the proposed approach has the most stable response and best accuracy.

\begin{figure}[h]
\centering
\includegraphics[trim=2 5 2 0,clip,width=.4\textwidth]{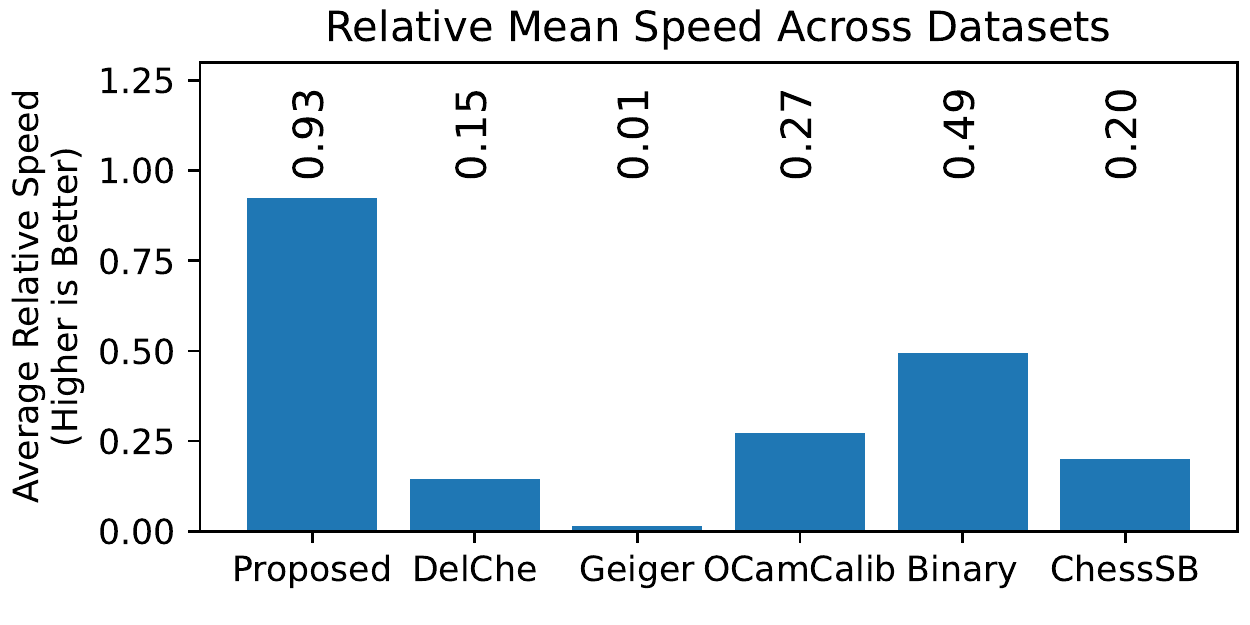}
\caption{Average mean relative runtime across all scenarios. A score of 1 would mean it is the top performer in every category.}
\label{fig:runtime-overall}
\end{figure}

The proposed library is 1.9 times faster on average than the second fastest and over 90 times faster than the second most accurate library, Figure \ref{fig:runtime-overall}. Proposed has a stable runtime across scenarios Table \ref{table:results_proposed}. Average runtime metrics hide erratic performance (i.e. very fast or slow on similar images) exhibited by some libraries, making them undesirable for real-time applications. One explanation for erratic performance is poor handling of excessive false positives, especially in higher resolution images. See Figure \ref{fig:pyramid-processing} for an example of how images can be covered in corners.

\begin{table}[ht]
\setlength{\tabcolsep}{4pt}
\centering
\input{performance_boofcv.tex}
\caption{Summary of Proposed Performance across all Scenarios}
\label{table:results_proposed}
\end{table}

Proposed approach's detection, accuracy, and runtime performance in each scenario is shown in Table \ref{table:results_proposed}. MP stands for image megapixels. When compared against other libraries in individual scenarios \cite{2021_Abeles_Techreport}, the proposed is always a top performer. Making it the most consistent library.

\section{Conclusion}

A new chessboard detector is proposed that is designed to handle larger images while being fast, accurate, and robust. This is accomplished using a new x-corner detector, where for the first time, blur is detected and used to select the best level in scale-space for corner location, dynamically adjust sample points for edge validation, and determine corner connectivity. Several incremental improvements are also proposed, such as correct handling of x-corner symmetry and n-best edge score. Performance is validated across a large set of scenarios using hand labeled and simulated data in what might be the most challenging chessboard dataset yet. Only the proposed detector is a top performer in both overall and all individual scenarios (particularly in larger degraded images), while being the fastest library by a large margin.





\bibliographystyle{IEEEtran}
\bibliography{mybib}

\end{document}

%% file: accuracy_table.tex
\begin{tabular}{| l | c c c | c c c | c c c | c c c |  }
\hline 
Library & \multicolumn{3}{c|}{ Sloppy }& \multicolumn{3}{c|}{ ThetaS }& \multicolumn{3}{c|}{ Motion }& \multicolumn{3}{c|}{ Stef2012 } \\
        & \tiny TP & \tiny E50 & \tiny E100 & \tiny TP & \tiny E50 & \tiny E100 & \tiny TP & \tiny E50 & \tiny E100 & \tiny TP & \tiny E50 & \tiny E100  \\
\hline 
Proposed & \tiny \textbf{13} & \tiny \textbf{0.42} & \tiny 2.17& \tiny \textbf{11} & \tiny 0.25 & \tiny 0.89& \tiny \textbf{37} & \tiny 0.67 & \tiny 4.45& \tiny \textbf{30} & \tiny \textbf{0.18} & \tiny 0.83\\
DelChe & \tiny \textbf{13} & \tiny 0.47 & \tiny \textbf{2.12}& \tiny \textbf{11} & \tiny 0.26 & \tiny \textbf{0.80}& \tiny 10 & \tiny \textbf{0.43} & \tiny \textbf{2.18}& \tiny 23 & \tiny 0.24 & \tiny 1.15\\
Geiger & \tiny \textbf{13} & \tiny \textbf{0.42} & \tiny 2.97& \tiny \textbf{11} & \tiny \textbf{0.24} & \tiny 0.94& \tiny 14 & \tiny 0.60 & \tiny 3.23& \tiny \textbf{30} & \tiny 0.19 & \tiny \textbf{0.81}\\
OCamCalib & \tiny 10 & \tiny 0.60 & \tiny 3.28& \tiny 6 & \tiny 0.44 & \tiny 1.44& \tiny 32 & \tiny 0.84 & \tiny 4.78& \tiny 26 & \tiny 0.36 & \tiny 4.28\\
Binary & \tiny 11 & \tiny 0.44 & \tiny 2.87& \tiny 9 & \tiny \textbf{0.24} & \tiny 4.86& \tiny 24 & \tiny 0.64 & \tiny 4.89& \tiny 14 & \tiny 0.20 & \tiny 2.94\\
ChessSB & \tiny \textbf{13} & \tiny 0.53 & \tiny 2.47& \tiny \textbf{11} & \tiny 0.31 & \tiny 4.22& \tiny 33 & \tiny 0.67 & \tiny 4.25& \tiny 26 & \tiny 0.25 & \tiny 3.15\\
\hline
\end{tabular}

%% file: performance_boofcv.tex
\begin{tabular}{l | c | c | c c | c c | c c}
\hline \hline
Dataset & MP & N & FP & FN & E50  & E100 & R50  & R100 \\
        &    &   &    &    & (px) & (px) & (ms) & (ms) \\
\hline \hline
border & 12.2 & 16 & 0 & 0 & 0.29 & 1.71 & 103 & 151 \\
caltech edges & 0.3 & 20 & 0 & 1 & 0.26 & 0.89 & 11.9 & 15.7 \\
challenge & 1.4 & 85 & 1 & 20 & 0.23 & 4.51 & 45.3 & 66.9 \\
close & 0.6 & 12 & 0 & 0 & 0.42 & 1.40 & 6.7 & 8.7 \\
kaist fisheye & 1.9 & 58 & 0 & 3 & 0.21 & 1.08 & 26.4 & 41.6 \\
distance angle & 0.5 &  6 & 0 & 0 & 0.21 & 0.59 & 10.6 & 30.8 \\
distance straight & 0.5 &  8 & 0 & 0 & 0.18 & 0.67 & 10.0 & 16.6 \\
distant & 0.5 &  7 & 0 & 0 & 0.20 & 0.63 & 16.4 & 19.7 \\
focus & 0.5 & 36 & 0 & 0 & 0.44 & 2.89 & 7.2 & 17.2 \\
focus large & 12.2 & 16 & 0 & 0 & 1.89 & 13.87 & 118 & 216 \\
gaus fisheye & 0.5 & 12 & 1 & 3 & 0.18 & 1.61 & 5.5 & 7.3 \\
gaus perfect & 0.1 & 12 & 0 & 0 & 0.02 & 0.10 & 2.3 & 3.2 \\
large & 10.0 &  6 & 0 & 0 & 0.31 & 1.25 & 180 & 232 \\
large shadow & 12.2 & 23 & 0 & 0 & 0.42 & 3.77 & 127 & 605 \\
motion blur & 0.3 & 38 & 1 & 0 & 0.67 & 4.45 & 5.4 & 7.5 \\
ocam fisheye190 & 0.4 &  8 & 0 & 1 & 0.22 & 1.16 & 8.3 & 13.0 \\
ocam kaidan omni & 0.3 & 17 & 0 & 1 & 0.26 & 1.03 & 10.3 & 15.2 \\
ocam ladybug & 0.8 & 13 & 0 & 0 & 0.26 & 0.84 & 10.8 & 16.1 \\
ocam mini omni & 0.4 & 15 & 0 & 0 & 0.22 & 3.25 & 17.9 & 27.7 \\
ocam omni & 1.2 & 14 & 0 & 0 & 0.29 & 1.11 & 27.0 & 36.1 \\
perfect & 0.3 &  8 & 0 & 0 & 0.00 & 0.01 & 7.5 & 11.0 \\
theta 5 & 0.5 & 11 & 0 & 0 & 0.25 & 0.89 & 13.0 & 16.0 \\
theta V & 3.7 & 46 & 0 & 0 & 0.32 & 2.59 & 88.2 & 134 \\
rotation flat & 0.5 & 12 & 0 & 0 & 0.28 & 0.97 & 17.1 & 20.5 \\
rotation vertical & 0.5 &  7 & 0 & 0 & 0.24 & 0.66 & 9.8 & 15.8 \\
shadow & 0.5 & 15 & 0 & 0 & 0.22 & 0.63 & 16.7 & 28.0 \\
sloppy13x10 & 2.1 & 13 & 0 & 0 & 0.42 & 2.17 & 34.1 & 43.6 \\
DSC-HX5V & 0.3 & 13 & 0 & 0 & 0.31 & 0.85 & 4.3 & 5.3 \\
stefano 2012 & 0.3 & 30 & 0 & 0 & 0.18 & 0.83 & 10.2 & 13.8 \\
\hline
\end{tabular}